\documentclass[runningheads,a4paper]{llncs}

\newcommand{\CurrentPath}{./Core}

\usepackage[T1]{fontenc}
\usepackage{hyperref} 
\usepackage{cite}
\usepackage{amsmath,amssymb,amsfonts}
\usepackage{algorithmic}
\usepackage{textcomp}
\usepackage{xcolor}
\usepackage[export]{adjustbox}

\usepackage{subfigure}

\usepackage{graphicx}
\usepackage{booktabs}

\usepackage{commath}
\usepackage{mathtools}

\usepackage[toc,page]{appendix}

\usepackage{ifthen} 
\usepackage{xurl}

\newboolean{IncludeAppendix}
\setboolean{IncludeAppendix}{true}

\newcommand{\CiteAsText}[1]{\cite{#1}}
\newcommand{\CiteAsRef}[1]{\cite{#1}}

\newcommand{\IsDecorator}[1]{\mathit{IsDec(#1)}}
\newcommand{\IsParallel}[1]{\mathit{IsPar(#1)}}
\newcommand{\IsParallelSynch}[1]{\mathit{IsParSynch(#1)}}

\newcommand{\IsRoot}[1]{\mathit{IsRoot(#1)}}
\newcommand{\IsSelector}[1]{\mathit{IsSel(#1)}}
\newcommand{\IsSelectorWithMemory}[1]{\mathit{IsSelWM(#1)}}
\newcommand{\IsSequence}[1]{\mathit{IsSeq(#1)}}
\newcommand{\IsSequenceWithMemory}[1]{\mathit{IsSeqWM(#1)}}
\newcommand{\IsSkipped}[2]{\mathit{Skipped(#1, #2)}}
\newcommand{\IsLeaf}[1]{\mathit{IsLeaf(#1)}}

\newcommand{\LastChild}[1]{\mathit{LChl(#1)}}
\newcommand{\FirstChild}[1]{\mathit{FChl(#1)}}
\newcommand{\Parent}[1]{\mathit{parent(#1)}}
\newcommand{\LNeigh}[1]{\mathit{lNeigh(#1)}}
\newcommand{\RNeigh}[1]{\mathit{rNeigh(#1)}}

\newcommand{\ActiveNode}[1]{\mathit{ActNode(#1)}}
\newcommand{\Active}[2]{\mathit{IsActive(#1, #2)}}
\newcommand{\Status}[2]{\mathit{status(#1, #2)}}
\newcommand{\NextNode}[3]{\mathit{NextNode(#1, #2, #3)}}
\newcommand{\NextNonSkippedChild}[2]{\mathit{Unskipped(#1, #2)}}
\newcommand{\ResumeFrom}[2]{\mathit{ResFrom(#1, #2)}}
\newcommand{\PriorSuccess}[2]{\mathit{PriorSucc(#1, #2)}}
\newcommand{\PriorFailure}[2]{\mathit{PriorFail(#1, #2)}}
\newcommand{\SkipChild}[2]{\mathit{SkipChl(#1, #2)}}

\newcommand{\Children}[1]{\mathit{Chl(#1)}}
\newcommand{\Ancestors}[1]{\mathit{Anc(#1)}}

\newcommand{\ParallelStatus}[2]{\mathit{ParStatus(#1, #2)}}
\newcommand{\FoundFailure}[2]{\mathit{IsFailure(#1, #2)}}
\newcommand{\SuccessCount}[2]{\mathit{NumSucc(#1, #2)}}
\newcommand{\RunningCount}[2]{\mathit{NumRun(#1, #2)}}
\newcommand{\SuccessThreshold}[1]{\mathit{SuccThresh(#1)}}
\newcommand{\SelectorStatus}[2]{\mathit{SelStatus(#1, #2)}}
\newcommand{\SequenceStatus}[2]{\mathit{SeqStatus(#1, #2)}}
\newcommand{\LeafStatus}[2]{\mathit{LeafStatus(#1, #2)}}
\newcommand{\DecoratorStatus}[2]{\mathit{DecStatus(#1, #2)}}

\newcommand{\Invalid}{I}
\newcommand{\Failure}{F}
\newcommand{\Running}{R}
\newcommand{\Success}{S}
\newcommand{\RootNode}{Root}

\begin{document}

\title{BehaVerify: Verifying Temporal Logic Specifications for Behavior Trees}
  %
  %
  \author{Serena S. Serbinowska\inst{1}\orcidID{0000-0002-9259-1586} \and\\
    Taylor T. Johnson\inst{1}\orcidID{0000-0001-8021-9923}}
  \authorrunning{S. Serbinowska, T. Johnson}
  %
  \institute{Vanderbilt University, Nashville TN 37235, USA\\
    \email{\{serena.serbinowska,taylor.johnson\}@vanderbilt.edu}}
  \maketitle              
  \begin{abstract}
    Behavior Trees, which originated in video games as a method for controlling NPCs but have since gained traction within the robotics community, are a framework for describing the execution of a task. BehaVerify is a tool that creates a nuXmv model from a PyTree. For composite nodes, which are standardized, this process is automatic and requires no additional user input. A wide variety of leaf nodes are automatically supported and require no additional user input, but customized leaf nodes will require additional user input to be correctly modeled. BehaVerify can provide a template to make this easier. BehaVerify is able to create a nuXmv model with over 100 nodes and nuXmv was able to verify various non-trivial LTL properties on this model, both directly and via counterexample. The model in question features parallel nodes, selector, and sequence nodes. A comparison with models based on BTCompiler indicates that the models created by BehaVerify perform better.
    \keywords{Behavior Tree \and Model Verification}
  \end{abstract}

\section{Introduction}\label{section:2022SEFM_Introduction}

Behavior Trees are a framework for describing the execution of a task that originated in computer games as a method of controlling Non-Player Characters (NPCs), but have since expanded into the domain of robotics \CiteAsRef{colledanchise2018book,ogren2020IEEE_Robotics_and_Automation_Letters}. Behavior Trees are split into composite nodes that control the flow through the tree and leaf nodes which execute actions. Behavior Trees have a variety of strengths: they facilitate code re-use (nodes and sub-trees can easily be attached), their modular nature makes reasoning about them easier, and changing one region of a tree doesn't affect how other regions function \CiteAsRef{biggar2020IEEE_Robotics_and_Automation_Letters}. However, at present, tools to verify the correctness of a Behavior Tree are scarce. Therefore, we present BehaVerify, a tool for converting a PyTree into a .smv file which can be verified using nuXmv \CiteAsRef{nuXmv}.

\paragraph{Contributions.} We present BehaVerify, a tool that enables verification with Linear Temporal Logic (LTL) model checking that improves upon BTCompiler, the only previously existing tool for such a task, in terms of run time and in ease of use with respect to Blackboard variables. Specifically, we present an automatic method to perform the translation and encoding of behavior trees to nuXmv models, a description of this method in a publicly available software tool, a characterization of the verification performance of these different encodings and how they compare to the models created by BTCompiler, and apply the tool to verify key LTL specifications of a challenging robotics case study for an underwater robot used as a controller in an ongoing DARPA project. However, we first define what Behavior Trees are.

\subsection{Background}

A Behavior Tree (BT) is a rooted tree. Each node has a single parent, save for the root which has no parent. A BT does nothing until it receives a tick event, at which point the tick event propagates throughout the tree. Composite nodes serve to control the flow of execution, determining which children receive tick events. By contrast, Leaf nodes are either actions, such as Accelerate, or guard checks, such as GoingToSlow. Leaf nodes do not have children. Finally, decorator nodes are used to customize the output of their children without actually modifying the children themselves, allowing for greater re-usability. Usually, a Decorator node will have one child.

There are three types of composite nodes: Sequence, Selector, and Parallel. Sequence nodes execute a sequence of children. A Sequence node returns a value if a child returns Failure or Running or there are no more children to run. Sequences return Failure if any child returns Failure, Running if any child returns Running, and Success if every child returned Success. Selector nodes, also known as Fallback nodes \CiteAsRef{biggar2020IEEE_Robotics_and_Automation_Letters, PyTrees}, execute children in order of priority. A Selector node returns a value if a child returns Success or Running or there are no more children to run. Selectors return Success if any child returns Success, Running if any child returns Running, and Failure if every child returned Failure. 

Parallel nodes execute all their children regardless of what values are returned. At least three different definitions exist for parallel nodes. The first definition, found in \CiteAsText{PyTrees}, states that parallel nodes return Failure if any child returns Failure, Success if a satisfactory subset of children return Success, and Running otherwise. The second definition, found in \CiteAsText{colledanchise2019IROS}, \CiteAsText{colledanchise2018IROS}, and \CiteAsText{giunchiglia2019SMC} is similar, but states that parallel nodes return Success only if all children return Success. The third definition, found in \CiteAsText{biggar2022ACM_Transactions_of_Cyber_Physical_Systems}, \CiteAsText{tumova2014IROS}, \CiteAsText{ogren2012AIAA}, \CiteAsText{colledanchise2018book}, \CiteAsText{colledanchise2022IEEE_Transactions_on_Robotics}, and \CiteAsText{colledanchise2021IEEE_Robotics_and_Automation_Letters}, states that parallel nodes return Success if at least $m$ children return Success, Failure if $n-m+1$ children return Failure, and Running otherwise. Here $n$ is the number of children the parallel node has and $m$ is a node parameter. BehaVerify, the tool created alongside this paper, was designed for PyTrees and therefore utilizes the definition presented in \CiteAsText{PyTrees}.

In addition to these differences, Composite nodes can be further characterized into Nodes with Memory and Nodes without Memory. The above definitions describe Nodes without Memory. Nodes with Memory allow the composite nodes to remember what they previously returned and continue accordingly. Thus a Sequence with Memory will not start from its first child if it previously returned Running and will instead skip over each child that returned Success. Similarly, a Selector with Memory will skip over each child that returned Failure. A Parallel node with Memory will only rerun children that returned Running.

However, memory is also not standardized. In \CiteAsText{PyTrees}, Nodes with Memory `forget' if one of their ancestors returns Success or Failure. So, for instance, if a Sequence with Memory returns Running, but its Parallel node parent returns Success, the Sequence with Memory will not behave as though it returned Running. However, in section 1.3.2 of \CiteAsText{colledanchise2018book}, the authors state ``Control flow nodes with memory always remember whether a child has returned Success or Failure, avoiding the re-execution of the child until the whole Sequence or Fallback finishes in either Success or Failure'', and notably makes no mention of Parallel nodes with Memory. Finally, note that PyTrees supports Selector with and without Memory, Sequences with and without Memory, and both types of Parallel nodes. However, the Parallel nodes with Memory and without Memory are instead called Synchronized Parallel and Unsynchronized Parallel, respectively.

Decorator nodes are generally used to augment the output of a child. For instance, a RunningIsFailure decorator will cause an output of Running to be interpreted as Failure. As there are many decorators, we omit attempting to fully list or describe them here.

\begin{figure}
  \centering
  \begin{minipage}{1.5cm}
    \includegraphics[width=1.5cm]{\CurrentPath/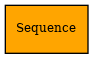}
  \end{minipage}
  \begin{minipage}{1.5cm}
    \includegraphics[width=1.5cm]{\CurrentPath/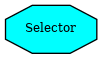}
  \end{minipage}
  \begin{minipage}{1.5cm}
    \includegraphics[width=1.5cm]{\CurrentPath/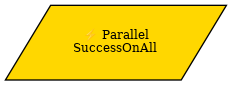}
  \end{minipage}
  \begin{minipage}{1.5cm}
    \includegraphics[width=1.5cm]{\CurrentPath/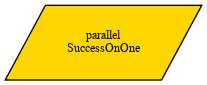}
  \end{minipage}
  \caption{Composite Nodes in PyTrees.} 
  \label{figure:2022SEFM_nodeTypes}
\end{figure}

Furthermore, we note that in many of the above works, Selector nodes are represented using ?, Sequence nodes are represented using $\rightarrow$, and Parallel nodes are represented using $\rightrightarrows$. However, we will utilize the notation given in PyTrees, as seen in \figref{figure:2022SEFM_nodeTypes}.

\subsection{The Blackboard}

In certain situations, such as when multiple nodes need to use the result of a computation, it can be useful to read and write information in a centralized location. This sort of shared memory is frequently called a Blackboard \CiteAsRef{PyTrees,broder2014unreal,unrealengine,cryengine}. Unfortunately, there are also drawbacks to using Blackboards. As \CiteAsText{shoulson2011MIG} points out, Blackboards can make BTs difficult to understand and reduce sub-tree reuse. Ultimately, however, the fact remains that in many cases there are substantial benefits to using a Blackboard, and various implementations, such as PyTrees seek to alleviate some of the aspects by creating visualization tools for blackboards \CiteAsRef{PyTrees}. Accordingly, BehaVerify supports Blackboard variables.
\section{Related Work}\label{section:Related_Work}

First, we clarify that the term ``Behavior Tree'' sometimes refer to different concepts. Behavior Trees exist as a formal graphical modeling language, as part of Behavior Engineering and are used for requirement handling \CiteAsRef{grunske2008Journal_of_Visual_Language_and_Computing}. These are not the BTs we are talking about. 

\subsection{Strengths and Uses of BTs}

In \CiteAsText{ogren2012AIAA}, the author shows how general Hybrid Dynamical Systems can be written as BTs and how this can be beneficial. Furthermore, the paper provides justifications for why BTs are useful to UAV guidance and control systems. \CiteAsText{biggar2021IEEE_Robotics_and_Automation_Letters} compares BTs to a variety of other Action Selection Mechanisms (ASM) and proves that unrestricted BTs have the same expressive capabilities as unrestricted Finite State Machines. \CiteAsText{biggar2020IEEE_Robotics_and_Automation_Letters} presents a framework for verifying the correctness of BTs without compromising on the main strengths of Behavior Trees, which they identify as modularity, flexibility, and re-usability.

\CiteAsText{ghzouli2020SLE} considers the various implementations of BTs, such as BehaviorTree.cpp and PyTrees, and examines a variety of repositories that utilize BTs. In \CiteAsText{tumova2014IROS} the authors propose an algorithm to translate an I/O automaton into a BT that connects high level planning and low level control. The authors of \CiteAsText{colledanchise2017IROS} demonstrate how it is possible to synthesize a BT that is guaranteed to be complete a task specified by LTL\@. This does require restricting LTL to a fragment of LTL, so there are limits to what BTs can be synthesized in this way. \CiteAsText{colledanchise2021arXiv} describes a tool-chain for designing, executing, and monitoring robots that uses BTs for controlling high level behaviors of the robots while \CiteAsText{colledanchise2021IROS} formalizes the context within which BTs are executed.

\subsection{Expanded BTs}

The capabilities of BTs have been expanded in several papers. In \CiteAsText{biggar2020arXiv}, the authors consider how it is possible to expand BTs by introducing K-BTs which replace Success and Failure with K different outputs. \CiteAsText{colledanchise2018IROS}, \CiteAsText{colledanchise2019IROS}, and \CiteAsText{colledanchise2022IEEE_Transactions_on_Robotics} introduce Concurrent BTs and expand on them by introducing various nodes designed to better enable synchronization in BTs that deal with concurrency. Meanwhile \CiteAsText{giunchiglia2019SMC} extends BTs to Conditional BTs, which enforce certain pre and post conditions on various nodes within the tree and introduces a tool which can confirm that the entire tree is capable of being executed based on the pre and post conditions given. \CiteAsText{safronov2020arXiv} extends BTs to Belief BTs which are better suited to dealing with non-deterministic outcomes of actions.

\subsection{Verification of BTs}

Some of the above works deal with the verification of BTs. \CiteAsText{biggar2020IEEE_Robotics_and_Automation_Letters}, for instance, presents an algorithm for the verification of BTs. \CiteAsText{colledanchise2017IROS}, on the other hand, presents a method by which to synthesize a BT that is guaranteed to be correct, thereby by-passing the need for verification, but the specifications are limited to a fragment of LTL\@. The only existing tool we were able to find that allows the user to create and verify LTL specifications for BTs is BTCompiler\footnote{\url{https://github.com/CARVE-ROBMOSYS/BTCompiler}}. Unfortunately, we were not able to install the tool, and as such our knowledge of it is somewhat limited. Most of what we understand comes from analyzing the various examples in the smv folder in the github repository.

From what we understand, BTCompiler uses the following assumptions and definitions. All composite nodes are assumed to have exactly 2 children. Parallel nodes do not have memory. Parallel nodes utilize the third definition presented in the background section. Sequence and Selector nodes with and without memory are supported. Unlike the implementation in PyTrees, nodes with memory do not `forget' if an ancestor terminates. Please note that the requirement that composite nodes have only 2 children does not impact expressiveness. By self-composing nodes, it is possible to effectively create a node with any number of children greater than 2. For a proof, see section 5.1 of \CiteAsText{colledanchise2018book}. Thus the only downside is potential model complexity and readability.

We will compare the models created by BTCompiler and BehaVerify.


\section{Overview of Approach}\label{section:2022SEFM_Overiew_of_Approach}

BehaVerify begins by recursively walking a PyTree and recording relevant information. This information includes what the type of each node is, recording any important parameters (like the Success policy for a parallel node), and the structure of the tree. Once this process has finished, BehaVerify begins to create the .smv file. Most of this process is straightforward. For instance, for each node type, BehaVerify creates a module (basically a class) in the .smv file. These modules are static and don't change between runs. For each node, BehaVerify creates an instance of a module with the necessary parameters, like what children the node has.

\begin{figure}
  \centering
  \includegraphics[width=2.3cm]{\CurrentPath/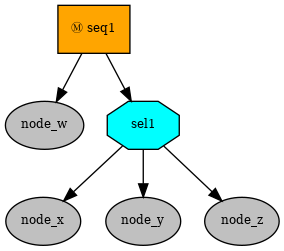}
  \caption{A simple BT}
  \label{figure:2022SEFM_ex1}
\end{figure}

However, not everything is simple or static. The primary sources of complexity are Nodes with Memory. A lazy approach to implementing Nodes with Memory is to have each node store an integer marking which child returned Running. Such an encoding can greatly increase the number of states in the model. Consider \figref{figure:2022SEFM_ex1}. Seq1 has two children, while sel1 has three. The lazy encoding would therefore produce six states to record which children returned Running. However, consider that if we know that node\_y returned Running, then sel1 will also return Running. Thus we only need four states.

Next, BehaVerify begins to handle the blackboard. BehaVerify has several ways of doing this. The first method is to have the user provide an input file which is simply included in the .smv file. Assuming no such file is provided, BehaVerify can generate the blackboard. If the user requests, the generated blackboard can be saved. This allows the user to modify the generated blackboard file and use it as an input file on subsequent runs. In addition, BehaVerify also allows the user to specify a file containing LTL specifications which are then included in the .smv file.

At this point, the .smv file is complete, and can be used with nuXmv \CiteAsRef{nuXmv}, either for simulation or verification.
\section{Encodings}\label{section:2022SEFM_Encodings}

BehaVerify uses two primary encodings: Leaf and Total. The general ideas behind these encodings are presented here. Note that the actual models BehaVerify creates for use with nuXmv differ from what is presented here, but the general motivations are the same. Also note that from this point forward, we write Success as $\Success$, Failure as $\Failure$, Running as $\Running$, and Invalid as $\Invalid$. For both encodings, it is useful to consider how a BT operates. A BT remains inactive until it receives a tick. Once a tick is received, it begins to propagate throughout the tree causing various nodes to execute. The path of the tick signal through the Tree is similar to a Depth First Search, though it will sometimes skip over branches of the tree. A basic version of the Leaf encoding explicitly follows the tick signal as it moves throughout the tree, tracing the exact path the tick signal takes through the tree. The Leaf encoding presented here includes some optimizations to improve performance, but the general idea is the same. The Total encoding doesn't follow the path of the signal. The state of the tree in the Total encoding is instead represented by a chain of dependencies and by considering the path of the tick signal through the tree, the chain can be resolved. Additional details follow.

\subsection{Leaf}

\begin{figure}
  \centering
  \includegraphics[width=4.5cm]{\CurrentPath/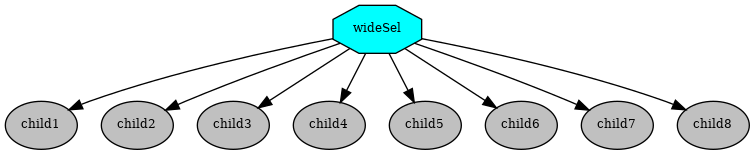}
  \caption{A selector node with many children.}\label{figure:2022SEFM_wideSel}
\end{figure}

As was mentioned, an intuitive encoding for BTs follows the path of the tick throughout the tree. At each time step $t$, one node is the Active Node ($\ActiveNode{t}$), its status is computed, and then another node becomes Active. Note that in this encoding each time step $t$ does NOT correspond to a tick. A tick instead occurs between any time steps $t$ and $t+1$ such that $\ActiveNode{t}=-1$. Now consider \figref{figure:2022SEFM_wideSel}. In this simple encoding, we would start at wideSel, then move to child1, then back to wideSel, then to child2, back to wideSel, etc., until one of the children returned $\Success$ or $\Running$, or we ran out of children. Thus this encoding spends many steps going through wideSel. The Leaf encoding realizes that the actual points of interest are the leaf nodes themselves. If child1 returns $\Success$ or $\Running$, then the tree returns a status. If child1 returns $\Failure$, then we need to check child2. Thus we can eliminate many unnecessary steps in the traversal of the tree by jumping from leaf to leaf. Formally, this encoding is as follows:

\begin{equation*}
  \ActiveNode{t+1} \coloneqq
  \begin{cases}
    \text{if } t+1 \leq 0 \text{, then } -1\\
    \text{else if } \ActiveNode{t} = -1 \text{, then } \NextNode{root}{t}{-1}\\
    \text{else } \NextNode{\ActiveNode{t}}{t}{\ActiveNode{t}}
  \end{cases}
\end{equation*}
So at each time step $t$, $\ActiveNode{t}$ either indicates a Node that is active or returns -1, which symbolizes the tree returning a value. In $\NextNode{n}{t}{prev}$, $n$ is either -1 or a node, $t$ is an integer indicating the time-step, and $prev$ is either -1 or a node and indicates which node asked for the Next Node. This is used to determine which node should be active next.  
\begin{equation*}
  \NextNode{n}{t}{prev} \coloneqq
\end{equation*}
\begin{equation*}
  \begin{cases}
    \text{if } n = -1 \text{, then} -1\\
    \text{else if } \Status{n}{t} \neq \Invalid \text{, then } \NextNode{\Parent{n}}{t}{n}\\
    \text{else if } \IsLeaf{n} \text{, then } n\\
    \text{else if } \IsSelector{n} \land prev = \Parent{n} \text{,}\\
    \quad\text{then }\NextNode{\NextNonSkippedChild{\FirstChild{n}}{t}}{t}{n}\\
    \text{else if } \IsSelector{n} \text{, then } \NextNode{\RNeigh{prev}}{t}{n}\\
    \text{else if } \IsSequence{n} \land prev = \Parent{n} \text{,}\\
    \quad\text{then }\NextNode{\NextNonSkippedChild{\FirstChild{n}}{t}}{t}{n}\\
    \text{else if } \IsSequence{n} \text{, then } \NextNode{\RNeigh{prev}}{t}{n}\\
    \text{else if } \IsParallel{n} \land prev = \Parent{n} \text{,}\\
    \quad\text{then }\NextNode{\NextNonSkippedChild{\FirstChild{n}}{t}}{t}{n}\\
    \text{else if } \IsParallel{n} \text{, then } \NextNode{\NextNonSkippedChild{prev}{t}}{t}{n}\\
    \text{else if } \IsDecorator{n} \land \SkipChild{n}{t} \text{, then } n\\
    \text{else } \NextNode{\FirstChild{n}}{t}{n}
  \end{cases}
\end{equation*}
$\Parent{\RootNode} = -1$ and otherwise $\Parent{n}$ returns the parent of $n$. \\ $\SkipChild{n}{t}$ returns True if at time $t$ decorator $n$ does not run it's child. $\IsLeaf{n}$, $\IsSelector{n}$, $\IsSequence{n}$, $\IsParallel{n}$, and $\IsDecorator{n}$ are all predicates that return True if the node $n$ is of the described type and False otherwise (all return False if $n=-1$). $\FirstChild{n}$ returns the first child of $n$, and $\RNeigh{n}$ indicates the right neighbor of $n$.
\begin{equation*}
  \NextNonSkippedChild{n}{t} \coloneqq
  \begin{cases}
    \text{if } \IsSkipped{n}{t} \text{, then } \NextNonSkippedChild{\RNeigh{n}}{t}\\
    \text{else } n
  \end{cases}
\end{equation*}
$\NextNonSkippedChild{n}{t}$ returns the first right Neighbor of $n$ that is not Skipped (Nodes with Memory can cause their children to be skipped in some cases). If there is no right neighbor, then $\RNeigh{n} = -1$.
\begin{equation*}
  \IsSkipped{n}{t} \coloneqq
  \begin{cases}
    \text{if } t \leq 0 \text{, then } \bot\\
    \text{else if } \exists a \in \Ancestors{n} \text{ s.t. } \Status{a}{t-1} \in \{\Success, \Failure\} \text{,}\\
    \quad\text{then } \bot\\
    \text{else if } \IsParallelSynch{\Parent{n}} \land \Status{n}{t-1} = \Success \text{,}\\
    \quad\text{then } \top\\
    \text{else if } \IsSequenceWithMemory{\Parent{n}} \land\\
    \quad\exists x \geq 1 \text{ s.t. } \Status{\RNeigh{n}^x}{t-1} = \Running \text{, then } \top\\
    \text{else if } \IsSelectorWithMemory{\Parent{n}} \land\\
    \quad\exists x \geq 1 \text{ s.t. } \Status{\RNeigh{n}^x}{t-1} = \Running \text{, then } \top\\
    \text{else } \IsSkipped{n}{t-1}
  \end{cases}
\end{equation*}
Here $\RNeigh{n}^x \coloneqq \RNeigh{\RNeigh{n}^{x-1}}$, with $\RNeigh{n}^1 \coloneqq \RNeigh{n}$. In other words, $\RNeigh{n}^x$ is the $x^{th}$ right neighbor.  $\Ancestors{n}$ is the set of nodes that are ancestors to $n$. This set does not include $n$ or $-1$. $\IsSequenceWithMemory{n}$ and $\IsSelectorWithMemory{n}$ check if $n$ is a Sequence/Selector node with memory, respectively.
\begin{equation*}
  \Status{n}{t} \coloneqq
\end{equation*}
\begin{equation*}
  \begin{cases}
    \text{if } \IsLeaf{n} \land \ActiveNode{t} = n \text{, then } \LeafStatus{n}{t}\\
    \text{else if } \IsSelector{n} \land (\exists c \in \Children{n} \text{ s.t. } \Status{c}{t} \in \{\Success, \Running\}) \text{, then } \Status{c}{t}\\
    \text{else if } \IsSelector{n} \land \Status{\LastChild{n}}{t} = \Failure \text{, then } \Failure\\
    \text{else if } \IsSequence{n} \land (\exists c \in \Children{n} \text{ s.t. } \Status{c}{t} \in \{\Failure, \Running\}) \text{, then } \Status{c}{t}\\
    \text{else if } \IsSequence{n} \land \Status{\LastChild{n}}{t} = \Success \text{, then } \Success\\
    \text{else if } \IsParallel{n} \land\\
    \quad(\exists c \in \Children{n} \text{ s.t. } (\Status{c}{t} \neq \Invalid) \land \NextNonSkippedChild{c}{t} = -1)\text{,}\\
    \quad\text{then } \ParallelStatus{n}{t}\\
    \text{else if } \IsDecorator{n} \land (\ActiveNode{t} = n \lor \Status{\FirstChild{n}}{t} \neq \Invalid) \text{,}\\
    \quad\text{then }\DecoratorStatus{n}{t}\\
    \text{else } \Invalid
  \end{cases}
\end{equation*}
$\Status{n}{t}$ describes the status of node $n$ at time step $t$. $\Children{n}$ is the set of children of $n$. If both $\IsDecorator{n}$ and $\ActiveNode{t} = n$, then $n$ is a decorator that skipped its child.
\begin{equation*}
  \ParallelStatus{n}{t} \coloneqq
  \begin{cases}
    \text{if } \FoundFailure{n}{t} \text{, then } \Failure\\
    \text{else if } \SuccessCount{n}{t} \geq \SuccessThreshold{n} \text{, then } \Success\\
    \text{else } \Running
  \end{cases}
\end{equation*}
\begin{equation*}
  \FoundFailure{n}{t} \coloneqq
  \begin{cases}
    \text{if } \exists a \in \Ancestors{n} \cup \{n\} \text{ s.t. } \Status{a}{t-1} \in \{\Success, \Failure\} \text{,}\\
    \quad\text{then } \bot\\
    \text{else } \FoundFailure{n}{t-1} \lor\\
    \quad\exists c \in \Children{n} \text{ s.t. } \Status{c}{t} = \Failure
  \end{cases}
\end{equation*}
\begin{equation*}
  \SuccessCount{n}{t} \coloneqq
  \begin{cases}
    \text{if } \exists a \in \Ancestors{n} \cup \{n\} \text{ s.t. } \Status{a}{t-1} \in \{\Success, \Failure\} \text{,}\\
    \quad\text{then } 0\\
    \text{else if } \exists c \in \Children{n} \text{ s.t. } \Status{n}{t} = \Success \text{,}\\
    \quad\text{then } \SuccessCount{n}{t-1} + 1\\
    \text{else } \SuccessCount{n}{t}
  \end{cases}
\end{equation*}

\subsection{Total}

Unlike the Leaf encoding,  in the Total encoding a tick occurs at each time step $t$ and we compute the entire state of the tree in one time step. Consider \figref{figure:2022SEFM_wideSel}. By definition, the status of wideSel is $\Success$ if a child returns $\Success$, $\Running$ if a child returns $\Running$, and $\Failure$ if all children return $\Failure$ (a status of $\Invalid$ is impossible for the root as the root will always run). The Total encoding uses this sort of definition directly for each node. Thus the status of each child is based on if the child runs and the custom code of the leaf node. As a result, in this case child3 will only run if child2 runs and returns $\Failure$, and child2 will only run if child1 runs and returns $\Failure$. This is all directly encoded, though it is done formulaically. The state of the tree is determined by resolving the dependency chain. Formally the encoding is defined as follows:

\begin{equation*}
  \Active{n}{t} \coloneqq
\end{equation*}
\begin{equation*}
  \begin{cases}
    \text{if } \IsRoot{n} \text{, then } \top\\
    \text{else if } \neg \Active{\Parent{n}}{t} \lor \IsSkipped{n}{t} \text{, then } \bot\\
    \text{else if } n = \FirstChild{\Parent{n}} \text{, then } \top\\
    \text{else if } \ResumeFrom{n}{t} \text{, then } \top\\
    \text{else if } \IsSelector{\Parent{n}} \text{, then } \Status{\LNeigh{n}}{t} = \Failure\\
    \text{else if } \IsSequence{\Parent{n}} \text{, then } \Status{\LNeigh{n}}{t} = \Success\\
    \text{else if } \IsParallel{\Parent{n}} \text{, then } \top\\
    \text{else } \bot
  \end{cases}
\end{equation*}
$\Active{n}{t}$ is True if at time $t$ node $n$ executed. In this encoding multiple nodes can be active at the same time. Notation is reused from the Leaf encoding where applicable. For instance, $\IsSelector{n}$ is defined as before. $\LNeigh{n}$ functions the same way as $\RNeigh{n}$, except with the Left Neighbor.
\begin{equation*}
  \IsSkipped{n}{t} \coloneqq
  \begin{cases}
    \text{if } t \leq 0 \text{, then } \bot\\
    \text{else if } \exists a \in \Ancestors{n} \text{ s.t. } \Status{a}{t-1} \in \{\Success, \Failure\} \text{,}\\
    \quad\text{then } \bot\\
    \text{else if } \IsParallelSynch{\Parent{n}} \land \Status{n}{t-1} = \Success \text{,}\\
    \quad\text{then } \top\\
    \text{else if } \IsSequenceWithMemory{\Parent{n}} \land\\
    \quad\exists x \geq 1 \text{ s.t. } \Status{\RNeigh{n}^x}{t-1} = \Running \text{, then } \top\\
    \text{else if } \IsSelectorWithMemory{\Parent{n}} \land\\
    \quad\exists x \geq 1 \text{ s.t. } \Status{\RNeigh{n}^x}{t-1} = \Running \text{, then } \top\\
    \text{else } \IsSkipped{n}{t-1}
  \end{cases}
\end{equation*}
$\IsSkipped{n}{t}$ is used to determine if a node with memory caused node $n$ to be skipped at time $t$.
\begin{equation*}
  \ResumeFrom{n}{t} \coloneqq \IsSequence{\Parent{n}} \land \exists x \geq 1 \text{ s.t. } \Status{\RNeigh{n}^x}{t-1} = \Running
\end{equation*}
Intuitively, $\ResumeFrom{n}{t}$ tells us if at time $t$ we are supposed to resume from node $n$ or not (only affects certain nodes with memory). As before $\Status{n}{t}$ is used to describe the status of a node $n$ at time $t$.
\begin{equation*}
  \Status{n}{t} \coloneqq
  \begin{cases}
    \text{if } \neg \Active{n}{t} \text{, then } \Invalid\\
    \text{else if } \IsSelector{n} \text{, then } \SelectorStatus{n}{t}\\
    \text{else if } \IsSequence{n} \text{, then } \SequenceStatus{n}{t}\\
    \text{else if } \IsParallel{n} \text{, then } \ParallelStatus{n}{t}\\
    \text{else if } \IsDecorator{n} \text{, then } \DecoratorStatus{n}{t}\\
    \text{else } \LeafStatus{n}{t}
  \end{cases}
\end{equation*}
\begin{equation*}
  \SelectorStatus{n}{t} \coloneqq
  \begin{cases}
    \text{if } \exists c \in \Children{n} \text{ s.t. } \Status{c}{t} \in \{\Success, \Running\} \text{,}\\
    \quad\text{then } \Status{c}{t}\\
    \text{else } \Failure
  \end{cases}
\end{equation*}
\begin{equation*}
  \SequenceStatus{n}{t} \coloneqq
  \begin{cases}
    \text{if } \exists c \in \Children{n} \text{ s.t. } \Status{c}{t} \in \{\Failure, \Running\} \text{,}\\
    \quad\text{then } \Status{c}{t}\\
    \text{else } \Success
  \end{cases}
\end{equation*}
\begin{equation*}
  \ParallelStatus{n}{t} \coloneqq
  \begin{cases}
    \text{if } \exists c \in \Children{n} \text{ s.t. } \Status{c}{t} = \Failure \text{, then } \Failure\\
    \text{else if } \SuccessCount{n}{t} \geq \SuccessThreshold{n} \text{, then } \Success\\ 
    \text{else } \Running
  \end{cases}
\end{equation*}
\begin{equation*}
  \SuccessCount{n}{t} \coloneqq \abs{\{c : c \in \Children{n} \land (\Status{c}{t} = \Success \lor \IsSkipped{c}{t})\}}
\end{equation*}
$\SuccessThreshold{n}$ represents the number of nodes that need to return Success for the parallel policy to return $\Success$. For the two default policies, Success On One and Success On All, the values would be $1$ and $\abs{\Children{n}}$ respectively. Therefore, if a node is a Parallel node and isn't $\Invalid$, then if any of the children returned $\Failure$ the node returns $\Failure$. Otherwise, it compares against the $\SuccessThreshold{n}$. $\SuccessCount{n}{t}$ is the number of children of $n$ that returned $\Success$ at time $t$. Since Leaf Nodes can be customized, it is impossible to fully characterize their behavior, and there are too many Decorator nodes to concisely list here. As such, we have $\DecoratorStatus{n}{t} \in \{\Success, \Failure, \Running\}$ and $\LeafStatus{n}{t} \in \{\Success, \Failure, \Running\}$. 

\subsection{BTCompiler}

The encoding for the BTCompiler, as best we understand it, has been included in \ifthenelse{\boolean{IncludeAppendix}}{Appendix~\ref{appendix:2022SEFM_BTCompiler_Encoding}}{\CiteAsText{serene2022SEFM_arXiv}}. Unfortunately, we were unable to install the tool. However, based on various examples in the BTCompiler repository, we concluded that the file `bt\_classic.smv'\footnote{\url{https://github.com/CARVE-ROBMOSYS/BTCompiler/blob/master/smv/bt_classic.smv}} contains the relevant encoding. The encoding presented in \ifthenelse{\boolean{IncludeAppendix}}{Appendix~\ref{appendix:2022SEFM_BTCompiler_Encoding}}{\CiteAsText{serene2022SEFM_arXiv}} is meant to approximate this, in the same way that the Leaf and Total encodings approximate the actual encodings used by BehaVerify.
\section{Results}\label{section:2022SEFM_Results}

We include the results of two main experiments: Checklist and BlueROV\@. Checklist is a parameterized example that takes as input an integer $n$ and produces a BT that contains $n$ checks which must either succeed or a fallback triggers. For each check we include two LTL specs, one to be proved and one to be disproved. Leaf\_v2, Total\_v2, Total\_v3, and BTC models were used in this experiment, where Leaf\_v2 is based on the Leaf encoding, Total\_v2 and Total\_v3 are based on the Total encoding, and BTC is based on the BTCompiler encoding. The other example is BlueROV, the controller in an ongoing DARPA project. As this example requires blackboard variables which BTCompiler does not support, it is not included, so only the 3 BehaVerify encodings are considered. We include timing results for verifying the LTL spec as well as memory usage. Timing values are based on nuXmv's `time' command. Maximum Resident Size values are based on nuXmv's usage command, which uses getrusage(2) \CiteAsRef{nuXmv}.  Maximum Resident Size is the maximum amount of RAM that is actually used by a process. All tests were run on a computer using Ubuntu 22.04 with 32 gb of ram and an i7{-}8700K Intel processor. Both the tool and instructions on how to recreate these tests are available\footnote{https://github.com/verivital/behaverify}. The tests only consider the time to verify LTL specifications in nuXmv. Time spent building the model in nuXmv is not included as it never exceeded .2 seconds. The time spent converting the BTs to models is not included as it is also fairly negligible, but can be found in \ifthenelse{\boolean{IncludeAppendix}}{Appendix~\ref{appendix:2022SEFM_BlueROV_smv_Timing_Results} and Appendix~\ref{appendix:2022SEFM_Checklist_smv_Timing_Results}}{\CiteAsText{serene2022SEFM_arXiv}}.

\subsection{Checklist and Parallel-Checklist}

\begin{figure}
  \centering
  \begin{minipage}[t]{5cm}
    \centering
    \includegraphics[width=5cm]{\CurrentPath/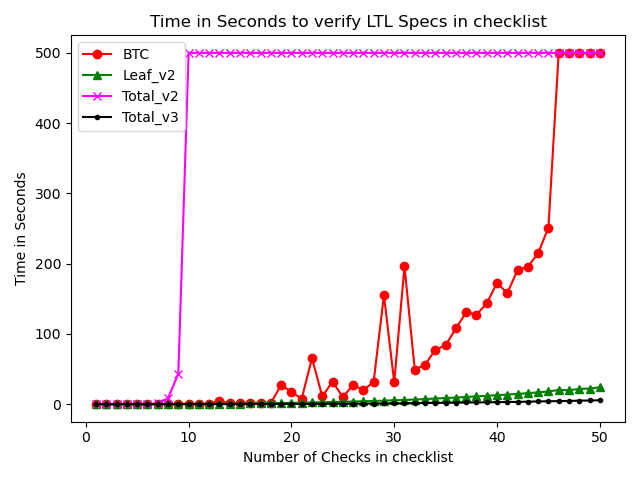}
  \end{minipage}
  \begin{minipage}[t]{5cm}
    \centering
    \includegraphics[width=5cm]{\CurrentPath/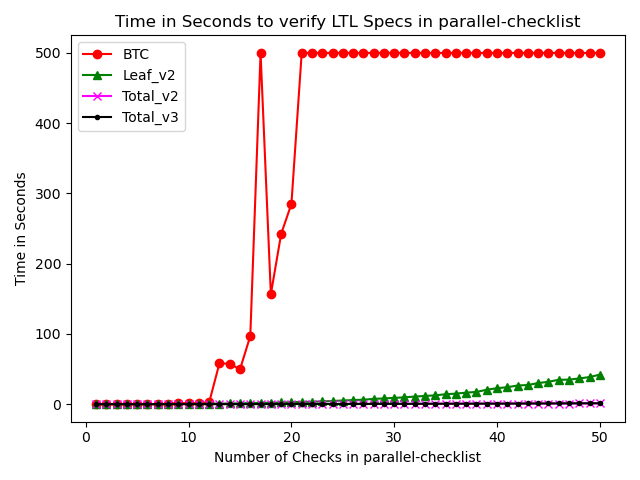}
  \end{minipage}
  \begin{minipage}[b]{5cm}
    \centering
    \includegraphics[width=5cm]{\CurrentPath/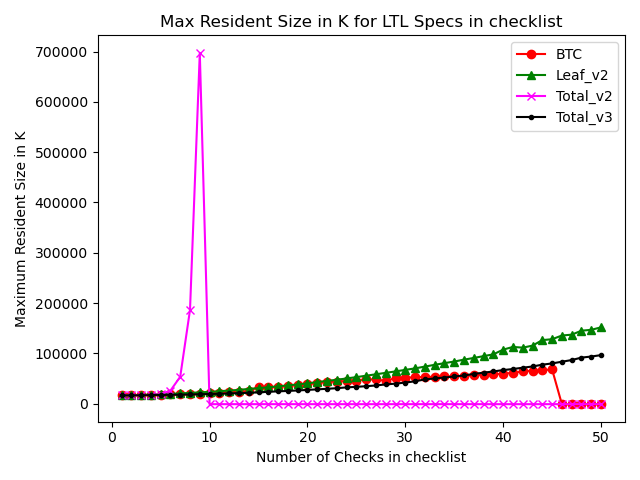}
  \end{minipage}
  \begin{minipage}[b]{5cm}
    \centering
    \includegraphics[width=5cm]{\CurrentPath/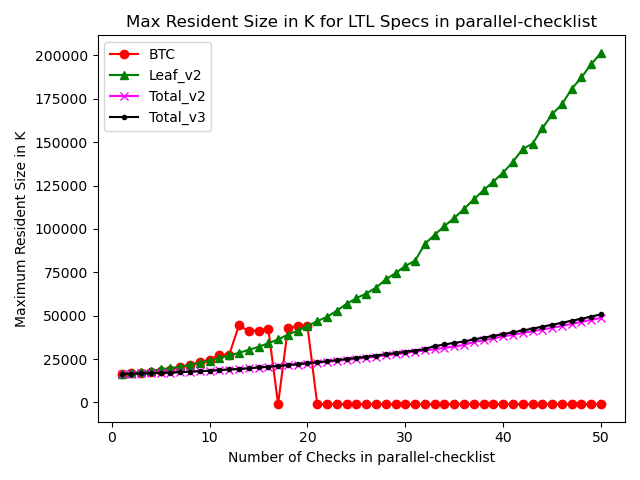}
  \end{minipage}
  \caption{Timing and memory results for verifying LTL specifications in nuXmv for Checklist and Parallel-Checklist. Timeout is set to 5 minutes. If a timeout occurred, a value of 350 is used for timing and -1000 for memory. After 3 timeouts, the remaining tests for the version are skipped. BTC is based on BTCompiler, Leaf\_v2 is a model based on the Leaf encoding, and Total\_v2 and Total\_v3 are models based on the Total encoding.}\label{figure:2022SEFM_checkTimeMem}
\end{figure}

The checklist examples consist of a series of checks that run in order by nested sequence nodes. Each check consists of a selector node, a safety check leaf node that can return $\Success$ or $\Failure$, and a backup node that can only return $\Success$. Thus if the safety check fails, the selector will run the backup which will return $\Success$. This process continues until each check has been run. See \ifthenelse{\boolean{IncludeAppendix}}{Appendix~\ref{appendix:2022SEFM_Checklist_Images}}{\CiteAsText{serene2022SEFM_arXiv}} for visual examples. Parallel-checklist replaces the sequence nodes with parallel nodes. Each check has two LTL specifications, one True and one False. The True/False specifications require that if a safety check fails, then a backup is triggered/not triggered. Due to differences in encodings, the specifications are slightly different for each version. We include one example here. The remainder can be found in \ifthenelse{\boolean{IncludeAppendix}}{Appendix~\ref{appendix:2022SEFM_Checklist_and_Parallel_Checklist_LTL_Specifications}}{\CiteAsText{serene2022SEFM_arXiv}}.

\begin{flalign*}
  &\text{For Total\_v2 and Total\_v3: }&\\
  &G (safety\_checkX.status = \Failure \implies backupX.status = \Success);&\\
  &G (safety\_checkX.status = \Failure \implies !(backupX.status = \Success));&
\end{flalign*}

\subsubsection{Checklist Results Discussion}

Having re-run the checklist and parallel checklist experiments three times for BTCompiler only, we have found that the spikes are present each time. These results can be found in \ifthenelse{\boolean{IncludeAppendix}}{Appendix~\ref{appendix:2022SEFM_BTCompiler_Results}}{\CiteAsText{serene2022SEFM_arXiv}}. The results are extremely similar, so we find it unlikely that this is a fluke. Furthermore, we note that there is a spike at 19 in both the checklist and parallel-checklist experiments. Since nuXmv is using a BDD model to verify the LTL Specifications, we assume that there is some sort of awkward break point with the number of variables that causes the efficiency to greatly suffer at certain points.

\begin{figure}
  \centering
  \begin{minipage}[t]{1.8cm}
    \centering
    \includegraphics[width=1.8cm]{\CurrentPath/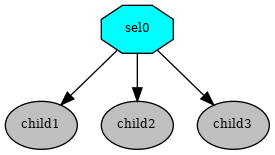}
  \end{minipage}
  \begin{minipage}[t]{1.8cm}
    \centering
    \includegraphics[width=1.8cm]{\CurrentPath/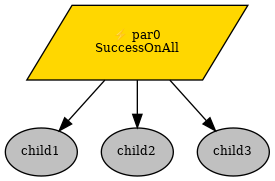}
  \end{minipage}
  \caption{Two examples with 3 children.}\label{figure:2022SEFM_3child}
\end{figure}

Note that Total\_v2 works much better on Parallel-Checklist than on Checklist. This is because of the logic chain created by Selector and Sequence nodes. Consider the Selector Example in \figref{figure:2022SEFM_3child}. The status of child3 depends on if child3 is active, which depends on the status of child2, which depends on if child2 is active, which depends on the status of child1, which depends on if child1 is active, which depends on if sel0 is active. The chain quickly becomes unmanageable (see \ifthenelse{\boolean{IncludeAppendix}}{Appendix~\ref{appendix:2022SEFM_Checklist_Images}}{\CiteAsText{serene2022SEFM_arXiv}} for visual examples of the BTs). This is not the case with Parallel-Checklist. Consider the Parallel Example in \figref{figure:2022SEFM_3child}. The status of child3 depends on if child3 is active, which depends only on par0 and what child3 returned last time. Thus the dependency chain is much shorter and thus Total\_v2 performs better on Parallel-Checklist. Total\_v3 avoid this by `guiding' nuXmv through this dependency chain by introducing intermediate variables.

Finally, note that the timing results in \figref{figure:2022SEFM_checkTimeMem} clearly demonstrate that the Total\_v3 encoding outperforms the rest.

\subsection{BlueROV}

\begin{table}
\centering
\caption{blueROV, Time in Seconds to Compute LTL}
\label{blueROV_LTL_time}
\begin{tabular}{lllll}
\toprule
Model & LTL Spec & Leaf\_v2 & Total\_v2 & Total\_v3 \\
\midrule
warnings only & low battery &    0.39 &     4.16 &     0.12 \\
warnings only & emergency stop &    0.48 &     4.21 &     0.14 \\
warnings only & home reached &    0.66 &        - &     1.70 \\
warnings only & obstacle &    0.54 &     7.79 &     0.17 \\
warnings only & sensor degradation &    0.49 &     4.11 &     0.13 \\
small & low battery &   23.43 &     5.06 &     0.33 \\
small & emergency stop &   30.47 &     6.40 &     1.02 \\
small & home reached &   31.48 &        - &     2.58 \\
small & obstacle &   39.34 &     9.87 &     0.39 \\
small & sensor degradation &   31.73 &     5.23 &     0.34 \\
full & low battery &   79.08 &     5.54 &     0.60 \\
full & emergency stop &  156.20 &     6.49 &     1.81 \\
full & home reached &  107.05 &        - &     3.59 \\
full & obstacle &  323.00 &    10.06 &     1.10 \\
full & sensor degradation &  106.16 &     6.57 &     1.46 \\
\bottomrule
\end{tabular}
\end{table}

We considered three versions of BlueROV\@: warnings only, small, and full. The differences between these versions is what range of values each blackboard variable is allowed to use. See \ifthenelse{\boolean{IncludeAppendix}}{Appendix~\ref{appendix:2022SEFM_BlueROV_Image}}{\CiteAsText{serene2022SEFM_arXiv}} for an image of the BT\@. We consider 5 sets of 2 LTL specifications. The timeout for each set of specifications was 10 minutes. For each warning, the first LTL specification requires that if the warning is set to True, then the appropriate Surface Task is triggered. This specification is False in all cases except battery low warning. The second LTL specification requires that if in a given tick a warning is set, then during that tick a surface task will trigger. This is true for all warnings except the home reached warning.

For the Leaf\_v2 encoding, these look as follows for battery:
\begin{flalign*}
  &G( next(battery\_low\_warning) = 1 \land active\_node = battery2bb \implies&\\
  &\qquad(active\_node > -1  U (active\_node = surface)));&\\
  &G( next(battery\_low\_warning) = 1 \land active\_node = battery2bb \implies&\\
  &\qquad(active\_node > -1  U (active\_node \in \{surface, surface1, surface2,&\\
  &\qquad\qquad surface3, surface4\})));&
\end{flalign*}
For the Total encodings, these look as follows for battery:
\begin{flalign*}
  &G( (next(battery\_low\_warning) = 1 \land battery2bb.active)&\\
  &\qquad\implies (surface.active));&\\
  &G((next(battery\_low\_warning) = 1 \land battery2bb.active)&\\
  &\qquad\implies (surface.active | surface1.active |&\\
  &\qquad surface2.active | surface3.active | surface4.active));&
\end{flalign*}

\subsubsection{BlueROV Results Discussion}

The BlueROV models differ from each other only in the number of values that each blackboard variable can take. Thus based on the results in Table 1, we can see that the Leaf\_v2 encoding has the worst scaling of the three with respect to blackboard variable size. Total\_v3 improves upon both Total\_v2 and Leaf\_v2. BTCompiler does not support blackboard variables.
\section{Conclusions and Future Work}\label{section:2022SEFM_Conclusions_and_Future_Work}

We introduced BehaVerify, a tool for turning a PyTree into a .smv file for use with nuXmv. We consider several possible encodings for this task and compared them to the encoding that BTCompiler uses. The results indicate that the encoding used by Total\_v3 is the best choice.

Future work includes general polish and improvements and expanding support for the various built-in nodes in PyTrees. In addition to this, we plan to re-work certain elements of BehaVerify. For instance, currently, in order for BehaVerify to detect blackboard variables in a PyTree using custom leaf nodes, the user must create a field that BehaVerify looks for within the custom node. This could certainly be handled better in the future. In terms of encodings, we plan to focus on Total\_v3. An improvement that has been considered, but not yet implemented, would be to restrict the incoming values to the leaf nodes to reduce state space. Specifically, in cases where a leaf node does not run, there is no need to consider the incoming status. Currently, this could be accomplished by tying the incoming value to the active value. However, this would likely cause worse performance for the same reason that Total\_v2 performs worse than Total\_v3. Therefore, the intended solution would be to, in some sense, enumerate all possible input values, which would hopefully shift some of the burden off of nuXmv and onto BehaVerify.
\section*{Acknowledgments}\label{section:2022SEFM_Acknowledgments}

The material presented in this paper is based upon work supported the Defense Advanced Research Projects Agency (DARPA) through contract number FA8750--18--C--0089, the Air Force Office of Scientific Research (AFOSR) award FA9550--22--1--0019, and the National Science Foundation (NSF) through grant number 2028001. Any opinions, findings, and conclusions or recommendations expressed in this publication are those of the authors and do not necessarily reflect the views of DARPA, AFOSR, or NSF\@.

\bibliographystyle{splncs04}
\bibliography{\CurrentPath/../Bibliography/Bibliography}

\ifthenelse{\boolean{IncludeAppendix}}{
  \appendix
  \newpage
\section{BTCompiler Encoding}\label{appendix:2022SEFM_BTCompiler_Encoding}

\begin{equation*}
  \Active{n}{t} \coloneqq
\end{equation*}
\begin{equation*}
  \begin{cases}
    \text{if } \IsRoot{n} \land t = 0 \text{, then } \top\\
    \text{else if } \IsRoot{n} \text{, then } \Status{n}{t-1} \neq \Invalid\\
    \text{else if } \IsSelector{\Parent{n}} \text{, then }\\
    \quad(n = \FirstChild{\Parent{n}} \land \Active{\Parent{n}}{t}) \lor\\
    \quad(\Status{\FirstChild{\Parent{n}}}{t} = \Failure) \lor\\
    \quad(\IsSelectorWithMemory{\Parent{n}} \land\\
    \qquad\PriorFailure{\FirstChild{\Parent{n}}}{t} \land\\
    \qquad\Active{\Parent{n}}{t})\\
    \text{else if } \IsSequence{\Parent{n}} \text{, then }\\
    \quad(n = \FirstChild{\Parent{n}} \land \neg \PriorSuccess{n}{t} \land \Active{\Parent{n}}{t}) \lor\\
    \quad(\Status{\FirstChild{\Parent{n}}}{t} = \Success \lor\\
    \quad(\IsSequenceWithMemory{\Parent{n}} \land\\
    \qquad\PriorSuccess{\FirstChild{\Parent{n}}}{t} \land\\
    \qquad\Active{\Parent{n}}{t}))\\
    \text{else if } \IsParallel{\Parent{n}} \text{, then }\\
    \quad(n = \FirstChild{\Parent{n}} \land \Active{\Parent{n}}{t}) \lor\\
    \quad(\Status{\FirstChild{\Parent{n}}}{t} \neq \Invalid)\\
    \text{else if } \IsDecorator{\Parent{n}} \text{, then } \Active{\Parent{n}}{t}
  \end{cases}
\end{equation*}

$\Active{n}{t}$ describes whether or not a node is active at time $t$. Multiple nodes can be active at the same time in this encoding. 

\begin{equation*}
  \PriorSuccess{n}{t+1} \coloneqq
  \begin{cases}
    \text{if } \Status{\LastChild{\Parent{n}}}{t+1} \in \{\Success, \Failure\} \text{, then } \bot\\
    \text{else if } \Status{n}{t+1} = \Success \text{, then } \top\\
    \text{else } \PriorSuccess{n}{t}
  \end{cases}
\end{equation*}

\begin{equation*}
  \PriorFailure{n}{t+1} \coloneqq
  \begin{cases}
    \text{if } \Status{\LastChild{\Parent{n}}}{t+1} \in \{\Success, \Failure\} \text{, then } \bot\\
    \text{else if } \Status{n}{t+1} = \Failure \text{, then } \top\\
    \text{else } \PriorFailure{n}{t}
  \end{cases}
\end{equation*}

$\PriorSuccess{n}{t}$ and $\PriorFailure{n}{t}$ are used to track where Nodes with Memory should begin from. 

\begin{equation*}
  \Status{n}{t} \coloneqq
\end{equation*}
\begin{equation*}
  \begin{cases}
    \text{if } \IsSelector{n} \land \Status{\FirstChild{n}}{t} \in \{\Running, \Success\} \text{, then } \Status{\FirstChild{n}}{t}\\
    \text{else if } \IsSelector{n} \text{, then } \Status{\LastChild{n}}{t}\\ 
    \text{else if } \IsSequence{n} \land \Status{\FirstChild{n}}{t} \in \{\Running, \Failure\} \text{, then } \Status{\FirstChild{n}}{t}\\
    \text{else if } \IsSequence{n} \text{, then } \Status{\LastChild{n}}{t}\\
    \text{else if } \IsParallel{n} \land \neg \Active{\LastChild{n}}{t} \text{, then } \Invalid\\
    \text{else if } \IsParallel{n} \land \SuccessCount{n}{t} \geq \SuccessThreshold{n} \text{, then } \Success\\
    \text{else if } \IsParallel{n} \land \SuccessThreshold{n} > \SuccessCount{n}{t} + \RunningCount{n}{t} \text{,}\\
    \quad\text{then } \Failure\\
    \text{else if } \IsParallel{n} \text{, then } \Running\\
    \text{else if } \IsDecorator{n} \text{, then } \DecoratorStatus{n}{t}\\
    \text{else if } \neg \Active{n}{t} \text{, then } \Invalid\\
    \text{else } \Status{n}{t} \in \{\Success, \Running, \Failure\}
  \end{cases}
\end{equation*}

While there are obvious differences between this and the Leaf encoding, overall it is closer to the Leaf encoding than the Total encoding. As with the Leaf encoding, the BTCompiler encoding does not consider the entire tree at the same time.
\newpage
\section{Checklist and Parallel-Checklist LTL Specifications}\label{appendix:2022SEFM_Checklist_and_Parallel_Checklist_LTL_Specifications}

\begin{flalign*}
  &\text{For BTCompiler: }&\\
  &G (safety\_checkX.status = \Failure \implies backupX.enable = TRUE);&\\
  &G (safety\_checkX.status = \Failure \implies backupX.enable = FALSE);&\\
  &\text{For Leaf\_v2: }&\\
  &G (safety\_checkX.status = \Failure \implies&\\
  &\qquad(!(active\_node = -1) U backupX.status = TRUE));&\\
  &G (safety\_checkX.status = \Failure \implies&\\
  &\qquad!(!(active\_node = -1) U backupX.status = TRUE));&\\
  &\text{For Total\_v2 and Total\_v3: }&\\
  &G (safety\_checkX.status = \Failure \implies backupX.status = \Success);&\\
  &G (safety\_checkX.status = \Failure \implies !(backupX.status = \Success));&
\end{flalign*}
\newpage
\section{BTC Results}\label{appendix:2022SEFM_BTCompiler_Results}

\begin{figure}
  \centering
  \begin{minipage}[t]{5cm}
    \centering
    \includegraphics[width=5cm]{\CurrentPath/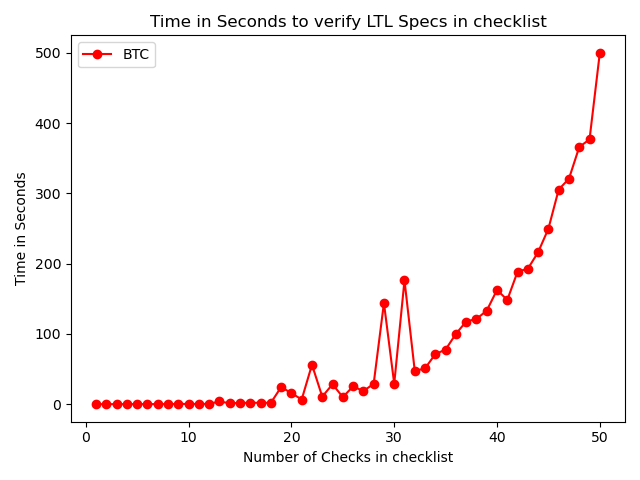}
  \end{minipage}
  \begin{minipage}[t]{5cm}
    \centering
    \includegraphics[width=5cm]{\CurrentPath/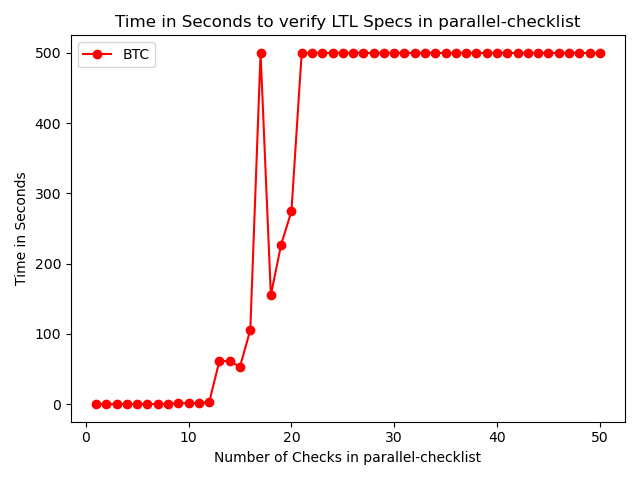}
  \end{minipage}
  \begin{minipage}[t]{10cm}
    \centering
    First Re-Run
  \end{minipage}
  \begin{minipage}[t]{5cm}
    \centering
    \includegraphics[width=5cm]{\CurrentPath/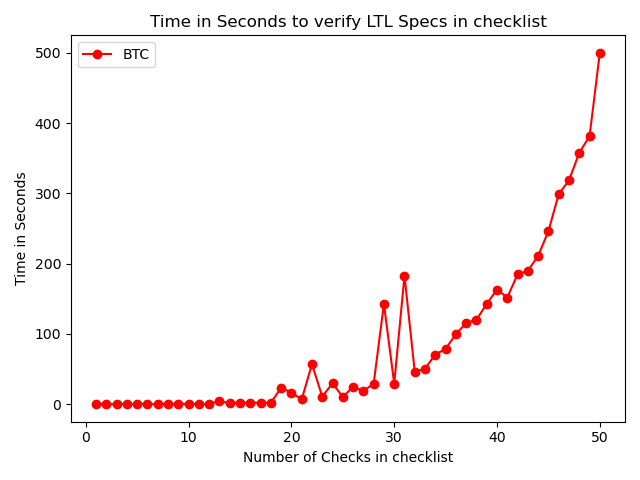}
  \end{minipage}
  \begin{minipage}[t]{5cm}
    \centering
    \includegraphics[width=5cm]{\CurrentPath/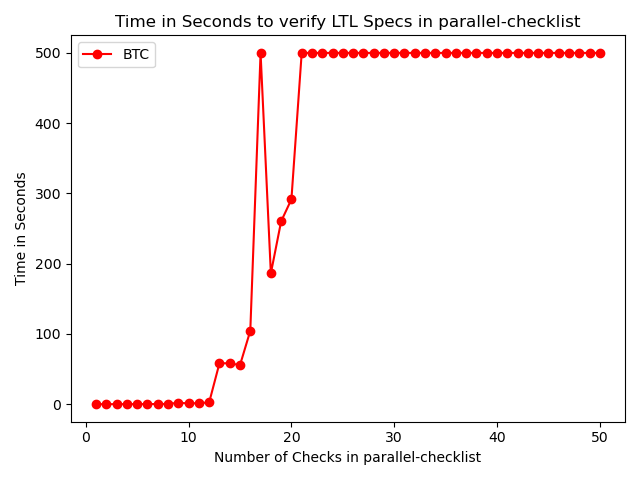}
  \end{minipage}
  \begin{minipage}[t]{10cm}
    \centering
    Second Re-Run
  \end{minipage}
  \begin{minipage}[t]{5cm}
    \centering
    \includegraphics[width=5cm]{\CurrentPath/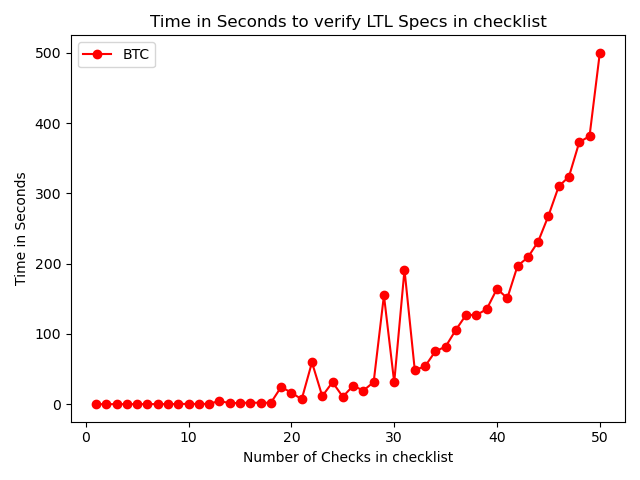}
  \end{minipage}
  \begin{minipage}[t]{5cm}
    \centering
    \includegraphics[width=5cm]{\CurrentPath/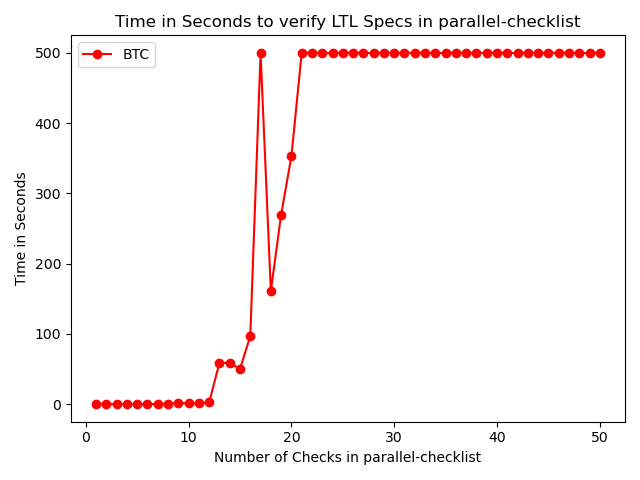}
  \end{minipage}
  \begin{minipage}[t]{10cm}
    \centering
    Third Re-Run
  \end{minipage}
  \caption{Anomaly Re-Runs for BTCompiler on Checklist and Parallel-Checklist. Timeout is set to 7 minutes. If a timeout occurred, a value of 500 seconds is used. If 5 timeouts occur, the remaining tests are automatically considered timeouts. In all three runs, the spikes occur in the same places, and overall all the graphs are very similar.}\label{figure:2022SEFM_anom}
\end{figure}

\newpage
\section{BlueROV .smv Timing Results}\label{appendix:2022SEFM_BlueROV_smv_Timing_Results}

\begin{table}
\centering
\caption{blueROV, Time in Seconds to Create .smv}\label{table:2022SEFM_BlueROV_build}
\begin{tabular}{lllll}
\toprule
Model  & Leaf\_v2 & Total\_v2 & Total\_v3 \\
\midrule
full          &    0.70 &     0.60 &     0.60 \\
small         &    0.60 &     0.60 &     0.60 \\
warnings only &    0.60 &     0.60 &     0.60 \\
\bottomrule
\end{tabular}
\end{table}

As is clear from Table~\ref{table:2022SEFM_BlueROV_build}, the amount of time taken to actually create the models is entirely negligible. Timing was done using memtime. Note that these results do not include the time it takes to print the model to a terminal or a file, as the output was instead sent to /dev/null.
\newpage
\section{Checklist .smv Timing Results}\label{appendix:2022SEFM_Checklist_smv_Timing_Results}

\begin{figure}
  \centering
  \begin{minipage}[t]{5cm}
    \centering
    \includegraphics[width=5cm]{\CurrentPath/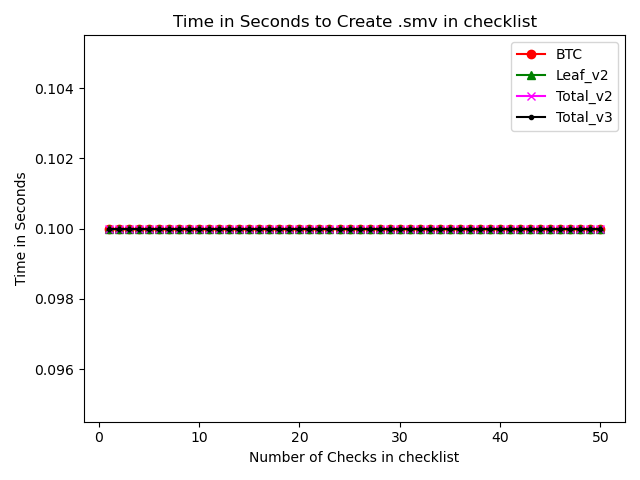}
  \end{minipage}
  \begin{minipage}[t]{5cm}
    \centering
    \includegraphics[width=5cm]{\CurrentPath/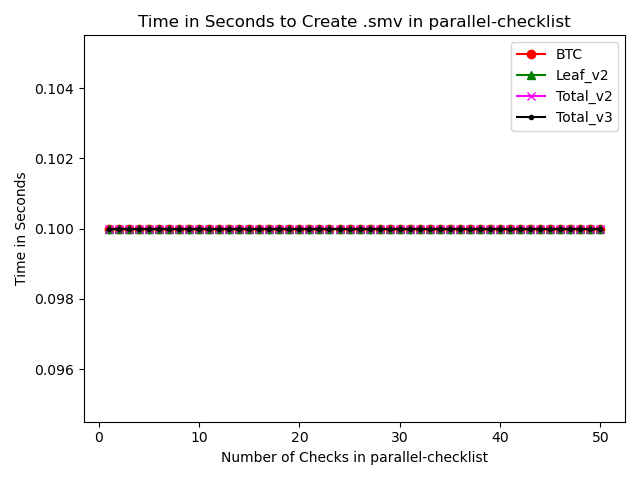}
  \end{minipage}
  \caption{Time in seconds for BehaVerify and BTCompiler to create .smv files for Checklist (left) and Parallel-Checklist (right). Note that the BTCompiler results are not based on the original tool, but rather on our recreation of it, as we could not install the original.}\label{fig:pointless}
\end{figure}

As is clear from \figref{fig:pointless}, the amount of time taken to actually create the models is entirely negligible. Timing was done using memtime. Note that these results do not include the time it takes to print the model to a terminal or a file, as the output was instead sent to /dev/null.
\newpage
\section{Checklist Images}\label{appendix:2022SEFM_Checklist_Images}

\begin{figure}
  \centering
  \begin{minipage}[t]{3cm}
    \centering
    {\includegraphics[width=3cm]{\CurrentPath/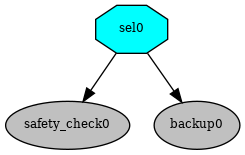} }
  \end{minipage}
  \begin{minipage}[t]{5cm}
    \centering
    {\includegraphics[width=5cm]{\CurrentPath/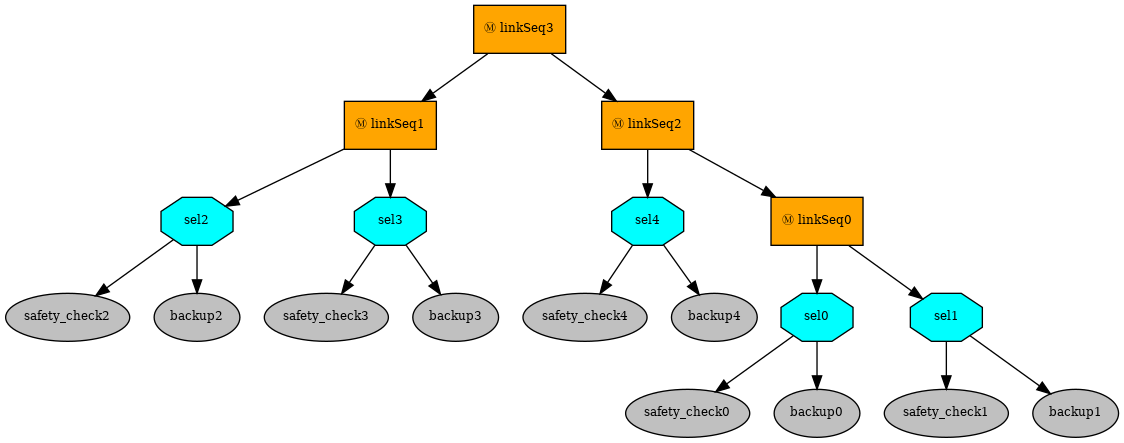} }
  \end{minipage}
  \caption{Checklist 1 and 5.}\label{figure:2022SEFM_checklist}
\end{figure}
\newpage
\section{BlueROV Image}\label{appendix:2022SEFM_BlueROV_Image}

\begin{figure}
  \centering
  \includegraphics[width=2.5cm]{\CurrentPath/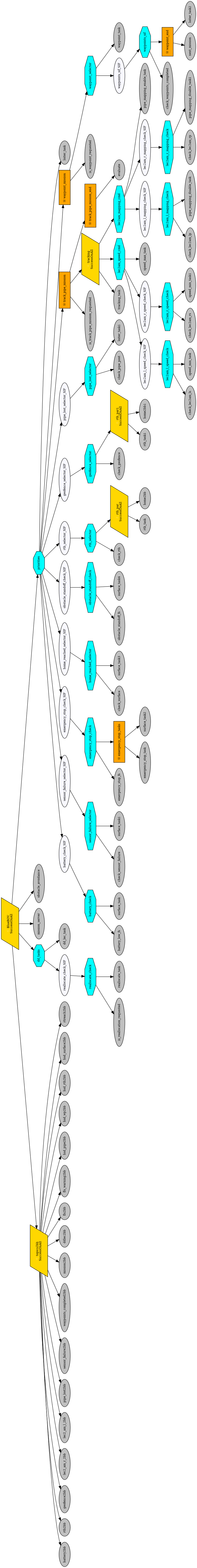}
  \caption{BT for BlueROV, which controls a UUV.}\label{figure:2022SEFM_BlueROV}
\end{figure}

}{}

\end{document}